\documentclass[english]{article}
\usepackage[T1]{fontenc}
\usepackage[latin9]{inputenc}
\usepackage{amsthm}
\usepackage{amsmath}
\usepackage{graphicx}

\makeatletter
\usepackage{emcsr2010}

\makeatother

\usepackage{babel}
\begin{document}

\title{Agent-based Exploration of Wirings of\\ Biological Neural Networks:
Position Paper}

\author{Önder Gürcan, O\u{g}uz Dikenelli\\Department of Computer Engineering
\\Ege University\\Bornova, Izmir, Turkey\\{\small {onder.gurcan,oguz.dikenelli}@ege.edu.tr}\And
Kemal S. Türker\\
Center for Brain Research\\
Ege University\\Bornova, Izmir, Turkey\\
{\small turker.77@gmail.com}}
\maketitle
\begin{abstract}
The understanding of human central nervous system (CNS) depends on
knowledge of its wiring. However, there are still gaps in our understanding
of its wiring due to technical difficulties. While some information
is coming out from human experiments, medical research is lacking
of simulation models to put current findings together to obtain the
global picture and to predict hypotheses to lead future experiments.
Agent-based modeling and simulation (ABMS) is a strong candidate for
the simulation model. In this position paper, we discuss the current
status of ``neural wiring'' and ``ABMS in biological systems''.
In particular, the paper discusses that the ABMS context provides
features that require for exploration of wiring of biological neural
networks.
\end{abstract}

\section{Introduction}

The understanding of the brain and the central nervous system (CNS)
depends on knowledge of their wiring\cite{Turker2005}. Since direct
recording from individual human neurons is impossible, indirect methods
are used. In indirect methods, a particular group of nerve fibres
or cells is stimulated and the responses of neurons that are affected
by the stimulus (reflexes) are recorded. Up until 1994\cite{Turker1994},
the wiring of the human CNS was estimated by counting the number of
neuron discharges that occur at specific times following a stimulus.
But this indirect method is open to numerous methodological errors
both in the stimulation and in the recording processes. Error free
indirect estimation of wiring in the human CNS has recently been tested\cite{Turker1999,Turker2003,Turker2005}
and now used in Ege University labs to reassess previously \textquoteleft{}established\textquoteright{}
wiring in the CNS. 

Although, neuroscientists are performing various experiments to explore
wirings, there are still gaps in our understanding of CNS because
of technical difficulties. For example, direct stimulation of nerves
is very difficult in some regions, since they are located deep. And
yet, there is no satisfactory theory on how these unknown parts of
CNS operate. Therefore, neuroscientists rely upon the knowledge that
is obtained in animal studies. Apparently, there is a strong need
to predict the characteristics of these gaps in the knowledge by putting
together information that is available from both human and animal
experiments. And such a prediction can be done using computational
simulation techniques.

Agent-based modeling and simulation (ABMS) seems as a strong candidate
for the simulation work and hence the solution to the problem of putting
information together to predict hypotheses for future studies. ABMS
is a new approach to modelling systems and is composed of interacting,
autonomous agents\cite{Macal2006}. It is a powerful and flexible
tool for understanding complex adaptive systems such as biological
systems. 

Biological systems are highly robust, flexible and has ability to
adapt to the changing circumstances. They are composed of bio-entities
that operate in naturally dynamic environments. It is widely accepted
that ABMS coordinated by self-organization and emergence mechanisms
are an effective way to design biological systems\cite{Serugendo2005}.
Because it is possible to associate different elements of a biological
process to independent computing entities (agents)\cite{Amigoni2007}.
Furthermore, ABMS allows explicitly modeling the environment in which
bio-entities operate.\textbf{ }

Biological neural networks can be understood in terms of complex networks.
Characterizing structure and function of complex networks\cite{Strogatz2001,Albert2002,Boccaletti2006}
is an interdisciplinary approach called network science\cite{Borner2007}.
Recent collaborative studies in network science and neuroscience show
that CNS have features of complex networks - such as small-world topology,
highly connected hubs and modularity\cite{Bullmore2009}. In another
work, it has been shown that an initially random wiring diagram can
evolve to a functional state characterized by a small-world topology
of the most strongly connected nodes and by self-organized critical
dynamics\cite{Siri2007}. Thus, it seems that neural wiring problem
can be reduced to network formation problem in which each node has
discretion in forming its links in the network relationship. In this
sense, abstractions of neurodevelopmental mechanisms\cite{vanOoyen2003}
of biological nervous systems can be used to form networks.

We believe that the use of ABMS and self-organizing dynamics, along
with the adoption of neuroscience and network science knowledge will
make it possible to harness the complexity of this problem domain
by delegating software agents to simulate bio-entities. 

Additionally, we wish to use ABMS in neuroscience as an adjunct to
laboratory and theoretical research. Simulation can be seen as a substitute
for an experiment that is impossible to perform in reality, where
impossibility can be either theoretical or pragmatical\cite{Hartmann1996}.
Thus, an ABMS tool to investigate neural wiring can also be used to
make ``in-machina'' experiments and to test hypothesis. 

This paper is organized as follows. The next section explains the
wiring problem of biological neural networks. In Section 3 ABMS in
biology is discussed. Section 4 states how ABMS can be used for exploration
of neural wiring. Related work is given in Section 5 and finally,
Section \ref{sec:Conclusion} concludes the paper.

\section{\label{sec:Wirings-of-BNN}Wiring of Biological Neural Networks}

The primary aim of neuroscientist working on neural wiring research
is to investigate functional connections in between biological neural
networks. This way we will have a better understanding about how CNS
works. This primary aim can be broken into the following specific
aims:
\begin{itemize}
\item To produce satisfactory explanations for the current situation.

\begin{itemize}
\item To be able to summate information available from different resources
for different pathways. 
\item To be able to pinpoint gaps in the system as the final working of
the system is known. 
\end{itemize}
\item To be able to bring out testable hypotheses for the unknown sections
of the system. 
\end{itemize}
With its multiplicity of cell types and complex patterns of cellular
interactions, the nervous system represents the most complex organ
of animals. Understanding how neuronal circuits are wired is one of
the holy grails of neuroscience. Besides many experimental advances
in determining the cellular machinery, theoretical approaches have
also proven to be useful tools in analyzing this machinery. A quantitative
understanding of neural wiring can allow us to make predictions, generate
and test hypotheses, and appraise established concepts in a new light.

The nervous system is a network of specialized cells that communicate
information about an organism's surroundings and itself. It processes
this information and generates reactions in other parts of the body.
It is composed of neurons and other specialized cells called glial
cells that aid in the function of the neurons. A neuron is an excitable
cell in the nervous system that processes and transmits information
by electrochemical signalling. A typical neuron can be divided into
three functionally distinct parts, dendrites, soma and axon. Roughly
speaking, the dendrites play the role of the ``input device'' that
collect signals from other neurons and transmits them to the soma.
The soma is the ``central processing unit'' that performs an important
non-linear processing step (called ``integrate \& fire model''):
If the total input exceeds a certain threshold, then an output signal
(spike) is generated \cite{Gerstner2002}. The output signal is taken
over by the ``output device'', the axon, which delivers the signal
to other neurons. Furthermore, neurons respond to stimuli, and communicate
the presence of stimuli to the central nervous system, which processes
that information and sends responses to other parts of the body for
action.

A number of specialized types of neurons exist: sensory neurons respond
to touch, sound, light and numerous other stimuli affecting cells
of the sensory organs that then send signals to the spinal cord and
brain. Motor neurons receive signals from the brain and spinal cord
and cause muscle contractions and affect glands. Each sensory neuron
receives information from a special ending, receptor. There are many
receptors in the skin, in the muscle, in the joints and within the
viscera. 

\begin{figure}
\begin{centering}
\includegraphics[scale=0.22]{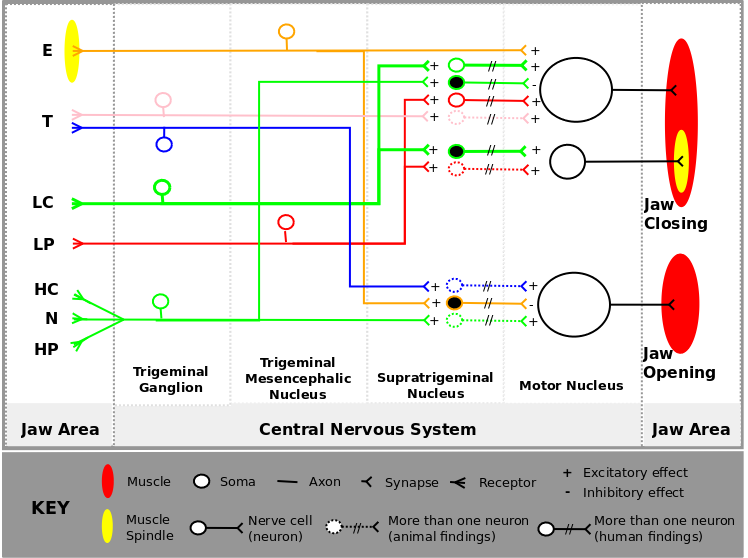}
\par\end{centering}

\caption{\label{fig:Wiring-diagram-of-Human-Masticatory-System}Wiring diagram
of human masticatory system representing pathways to jaw opening (JO)
and jaw closing (JC) muscles (different colors of neurons indicates
different pathways).}
\end{figure}

Figure \ref{fig:Wiring-diagram-of-Human-Masticatory-System} shows
a simplified wiring diagram of human masticatory system during no
mastication%
\footnote{The diagram is simplified due to space limitations.%
}. This diagram is a result of various experiments established in Ege
University labs\cite{Naser-Ud-Din2010,Lobbezoo2009,Sowman2008}, other
past experiments and animal studies (reviewed in detail in \cite{Turker2002}).\textit{
}The reason for using animal studies is: direct electrical stimulation
of nerves is very difficult in some parts of the jaws region, since
these nerves are located deep in the face and close to numerous blood
vessels\cite{Scutter1997}. Even though the dashed lined neurons are
represented like a single neuron in the figure, there may be a few
(oligosynaptic) or many (polysynaptic) neurons (each one connected
to several other ones) in that part of the wiring diagram. 

Using human reflex research%
\footnote{Reflex pathways contain much less neurons than cortical pathways. %
}, neuroscientists are carrying on finding pathways as much as possible,
but as mentioned above investigating some parts of the picture is
almost impossible to obtain using current techniques. Consequently,
there must be computational tools (and techniques) that allow combining
the current findings to predict the wiring of the pathways that are
impossible to obtain in human subjects.

\section{\label{sec:ABMS}Agent-based Modeling and Simulation in Biology}

Agent-based modeling and simulation (ABMS) has increasingly been adopted
as a suitable approach for analyzing complex systems and evaluating
theories and models of complex systems (especially for social or biological
systems). ABMS is used in a broad range of domains, including social
and economical simulation, biological systems, traffic and crowd simulation
and in other domains\cite{Macal2006}. 

Since ABMS allows biological systems to be decomposed into several
independent but interacting entities, usage of ABMS for biological
systems is widespread\cite{Amigoni2007,Merelli2007}. Each day, more
promising evaluations of ABMS for solving biological problems are
being developed (e.g.: \cite{Maniadakis2009,Folcik2007,Christley2007}). 

Basically, the reasons for using ABMS in biology are twofold\cite{Amigoni2007}:
\begin{itemize}
\item \textit{ABMS in biology can be used to support information gathering,
processing and integration}. In other words, ABMS can be used to summate
information gathered from various experiments and can help us to understand
biological processes.
\item \textit{ABMS can be used to simulate the behavior of biological systems}.
As Hartmann says\cite{Hartmann1996}: ``Simulation can be seen as
a substitute for an experiment impossible to make in reality, where
impossibility can be either theoretical or pragmatical''. Hence,
by simulating the behavior of biological systems it becomes possible
produce testable hypotheses for the unknown parts of the system.
\end{itemize}
Following there paragraphs explain the features provided by ABMS that
support these reasons.

Biological systems are self-organized in their nature\cite{Camazine2001}.
A system is said to be ``self-organizing'' if \textit{it is able
to reorganize itself by managing the relations between components,
either topological, structural or functional, upon environment perturbations
solely via the interaction of its components, with no requirement
of external forces}\cite{Gardelli2009}. Self-organization allows
us to reduce the complexity of problem by concerning not the overall
system, but the behaviors of individual agents. It is widely accepted
that ABMS is well suited for simulating self-organizing systems\cite{Macal2006}.
For instance, when simulating biological systems an agent is a good
abstraction for representing bio-entities which represents a global
phenomena when put together. 

Environment also plays an important role when simulating biological
systems. There are two roles of the environment when simulating biological
systems\cite{Klugl2004}. Firstly, simulation can be used for modelling
environment. Real bio-entities operate in naturally dynamic complex
biological environments. Thus, when simulating the behavior of biological
systems this dynamism should be explicitly modeled\cite{Helleboogh2007}.
Secondly, environment can be used for simulation. Since ABMS can be
seen as simulated multi-agent systems situated in a simulated environment,
in simulations the modelled environment should always be a first class
entity that is as carefully developed as the agents themselves\cite{Klugl2004}.
This is especially true for self-organizing multi-agent systems, as
the agents' environment guides the selection and self-organization
process. Furthermore, the study of biological systems needs experiments
to explore their behaviors. Similarly, simulation models of such systems
must be run many times to explore if they behave as expected. Eventually,
to be able to assess simulation results, data generated by the simulation
runs and data collected from experiments should be compared. In such
a situation, environment can be used as a regulator for sake of calibration
in order to obtain results that can be analyzed and compared to actual
data\cite{Bandini2006}. 

However, running an agent-based model is an easy task, but the analysis
is not\cite{Richiardi2006}. Even for simple scale simulations, we
must cope with vast parameter space of the model. Thus, parameters
should be tuned in order to find the optimal behavior of an agent
performs. This optimal behavior is going to influence the global behavior
of its collective. Nevertheless, even if supported by a reference
tool, the tuning process can be quite time-consuming. It is apparent
that theories and tools allowing automatic tuning of parameters are
needed. Recent studies address the problem of automatic tuning of
parameters of agent-based simulation models\cite{Bonjean2009,Montagna2009,Gardelli2009,Terano2007,Fehler2006}. 

Besides the aforementioned clear advantages, the main problem when
using ABMS in biology is the level of trust to the outcome obtained
using ABMS \cite{Amigoni2007}. In other words, the weak validation
of the results obtained makes ABMS hard to trust. The reason for that
is the lacking of a governing thoery. Moreover, experimental data
obtained are sparse and thus a direct comparison to the results obtained
in simulation is sometimes difficult. Overcoming this drawback requires
developments of aferomentioned calibration mechanisms and, closer
collaborations between biologists and computer scientists\cite{Fisher2007}.

\section{Using ABMS for Exploration of Neural Wiring}

ABMS can be effectively used to tackle exploration of neural wiring
problem due to its aforementioned advantages. In this section we propose
that the use of ABMS and self-organizing dynamics, along with the
adoption of network science and neuroscience knowledge would lead
us to highlight neural wiring problem.

\subsection{Proposed Approach}

There is considerable amount of knowledge about developmental neuroscience.
The study of neural development aims to describe the cellular basis
of brain development and to address the underlying mechanisms. The
science of studying neural development by computational and mathematical
modeling is relatively new\cite{vanOoyen2003}. Neural development
models are used to study the development of the nervous system at
different levels of organization and at different phases of development,
from molecule to system and from neurulation to cognition.

Neurodevelopmental processes can be broadly divided into two classes:
activity-independent mechanisms and activity-dependent mechanisms.
Activity-independent mechanisms are generally believed to occur as
hardwired processes determined by genetic programs played out within
individual neurons. These include differentiation, migration and axon
guidance to their initial target areas. These processes are thought
of as being independent of neural activity and sensory experience.
Once axons reach their target areas, activity-dependent mechanisms
come into play. Although synapse formation is an activity-independent
event, modification of synapses and synapse elimination requires neural
activity. Activity-independent mechanisms are said to be self-organizing
dynamics of neurons. There is no information at a higher level of
organization than the individual neuron, so all the organization in
central nervous system and brain is an \textquoteright{}emergent property\textquoteright{}
of the interaction of large numbers of individual neurons.

Besides, the small-world architecture have been found in several empirical
studies of brain networks in humans and other animals and it is shown
that CNS have features of complex networks - such as small-world topology,
highly connected hubs and modularity\cite{Bullmore2009}. More than
that, it has been shown that an initially random wiring diagram can
evolve to a functional state characterized by a small-world topology
of the most strongly connected nodes and by self-organized critical
dynamics\cite{Siri2007}. 

To this end, inspiring from the above information gathered from neuroscience
and network science studies we may develop an agent-based model. 

An important consideration when we use ABMS to simulate a system is
deciding the level of abstraction to model the system. We can describe
a system using different levels, aspects, or representations. As Prem\cite{Prem1993}
suggests, the level should be the one where the prediction of the
behavior of the system is easiest; in other words, where we need least
information to make predictions\cite{Shalizi2001}. In this sense,
in our biological neural network model, agents will represent bio-entities
at cell and tissue level: neuron agents, receptor agents, and muscle
agents. 

Neuron agents will implement the activity independent (self-organizing)
mechanisms of neural development. An apt learning algorithm will then
be applied to make the network evolve to a small-world topology. Meanwhile,
environment will control neural development using global parameters
and characteristics. Since activity dependent mechanisms require neural
activity, they will be modeled using experimental data obtained to
calibrate the model together with receptor agents and muscle agents.

\section{Related Work}

Schoenharl et. al.\cite{Schoenharl2005} developed a toolkit for computational
neuroscientists to explore developmental changes in biological neural
networks. This toolkit develops complex network topologies in neural
networks using pruning. However, details of the methodology used (e.g.
how the initial random network is constructed) and of simulation parameters
(e.g. how the threshold parameter for pruning is obtained) are not
clear. 

Mano et. al.\cite{Mano2005} presents an approach to self-organization
in a dynamic neural network by assembling cooperative neuro-agents
(CNA). The network is initialized with only unconnected CNAs and then
during a learning period, the network self-organizes. The network
of CNAs they presented is able to define criteria for adapting the
genotypic transfer function, node strengths, connectivity between
nodes, neuron proliferation, and even neuron deaths. 

Maniadakis et. al.\cite{Maniadakis2009} addresses the development
of brain-inspired models that will be embedded in robotic systems
to support their cognitive abilities. They introduce a novel agent-based
coevolutionary computational framework for modeling assemblies of
brain areas. They specifically employ self-organized agent structures
to represent brain areas. Moreover, they introduce a ``hierarchical
cooperative coevolutionary'' scheme that effectively specifies the
structural details of autonomous, yet cooperating system components.
However, this work focuses on brain slices rather than reflex pathways
and aims to improve cognitive capabilities of robotic systems. But,
introduced mechanisms may be used to support research efforts in the
field of psychology and neuroscience.

Apart from agent-based approaches there are \textit{biological neural
network simulators.} They are used primarily to simulate spiking neural
networks which are present in the biology to study their operation
and characteristics. In this group we can find sophisticated simulators
such as GENESIS\cite{Bower2002} and NEURON\cite{Hines1997}. They
are designed to provide biologically realistic models of electrical
and chemical signalling in neurons and network of neurons. They support
the simulation of complex neural models with a high level of detail
and accuracy. However, this research into electrical signalling ignores
the fascinating problem of wiring of biological neural networks.

\section{\label{sec:Conclusion}Conclusion and Prospects}

In this paper we have shown that neuroscientists working on neural
wiring has a strong need of computer scientists. Then we have provided
a critical outlook on agent-based modelling and simulation in solving
their problems. 

The literature shows that, day by day, ABMS is getting more mature
for biological systems. However, although ABMS is a pioneering and
powerful approach in biology, research into the design and use of
agent-based models is still in its infancy and requires closer collaborations
between biologists and computer scientists. The inherent differences
between biological and computational models, along with the difficulty
of obtaining precise biological data, make both approaches indispensable\cite{Fisher2007}. 

Up to now, we have established a ``core'' biological model involving
various simplifications and assumptions (Section 2). We proposed then
a preliminary agent-based simulation model (Section 4). Next step
will be enhancing, implementing and calibrating the proposed model.
We will then compare \textit{in silico} experiments with \textit{in
vitro} biological experiments. As a result of comparison we will either
adjust our computatinal model or develop new/improved biological experiments
to revise the biological model. This cycle will proceed until we get
satisfactory results.

We see this study as the first step for understanding neural circuits.
Within the scope of the goals of neural wiring research, we are planning
to develop an agent-based simulator that combines currently available
bits of data into a manageable format. So that the working of the
entire nervous system can be tested. The simulator will be used to
explore the wiring of, possibly dynamic, black box networks. This
will probably lead us produce new techniques to construct and test
network development inspiring from developmental neuroscience as well
as we will use existing network development techniques. Furthermore,
the developed simulator will be used for predicting future human reflex
findings in via putting forward workable hypotheses about human CNS.
Meanwhile, we will demonstrate the power of the developed simulator
through a series of case studies.

\section{Acknowledgements}

Supported by the FP6 GenderReflex project MEXT-CT-2006-040317 and
the TUBITAK - 107S029 - SBAG-3556 grant from the Turkish Scientific
and Technological Research Organization. 

\bibliographystyle{named}
\bibliography{ogurcan}

\end{document}